\documentclass[onecolumn,journal]{IEEEtran}
\usepackage{amsmath,amsfonts,amssymb,amsthm}
\usepackage{algorithmic}
\usepackage{algorithm}
\usepackage{array}
\usepackage{subfigure}
\usepackage{textcomp}
\usepackage{stfloats}
\usepackage{url}
\usepackage{verbatim}
\usepackage{graphicx}
\def\BibTeX{{\rm B\kern-.05em{\sc i\kern-.025em b}\kern-.08em
    T\kern-.1667em\lower.7ex\hbox{E}\kern-.125emX}}
\usepackage{balance}

\usepackage{cite}
\usepackage{makecell}
\usepackage{adjustbox}

\usepackage{booktabs} 
\usepackage{multirow}
\usepackage{xcolor}

\begin{document}

\title{Learning Hyperplane Tree: A Piecewise Linear and Fully Interpretable Decision-making Framework}

\author{Hongyi Li, Jun Xu,, and William Ward Armstrong
\thanks{
 Corresponding author: Jun Xu.}
\thanks{
Hongyi Li and Jun Xu are with the Department of Automation, Harbin Institute of Technology, Shenzhen, 518055, China (e-mail: xujunqgy@hit.edu.cn).
}
\thanks{William Ward Armstrong is with  the Department of Computing Science, University of Alberta, Edmonton, Canada.}
}

\markboth{}%
{How to Use the IEEEtran \LaTeX \ Templates}

\maketitle

\begin{abstract}
This paper introduces a novel tree-based model, Learning Hyperplane Tree (LHT), which outperforms state-of-the-art (SOTA) tree models for classification tasks on several public datasets. The structure of LHT is simple and efficient: it partitions the data using several hyperplanes to progressively distinguish between target and non-target class samples. Although the separation is not perfect at each stage, \textcolor{black}{LHT effectively improves the distinction} through successive partitions.
During testing, a sample is classified by \textcolor{black}{evaluating the hyperplanes}
defined in the branching blocks and traversing down the tree until it reaches the corresponding leaf block. \textcolor{black}{The class of the test sample is then determined using the piecewise linear membership function defined in the leaf blocks, which is derived through least-squares fitting and fuzzy logic.} LHT is highly transparent and interpretable—at each branching block, the contribution of each feature to the classification can be clearly observed.
\end{abstract}
\section{Introduction}

Deep learning has made significant progress in fields such as image and language \cite{padhiary2025convergence}. However, when dealing with tabular data, particularly in scenarios with small sample sizes and strong feature heterogeneity, decision tree models—especially techniques like XGBoost \cite{chen2016xgboost}—remain the state-of-the-art (SOTA) and continue to dominate data science competitions \cite{shwartz-ziv2021tabular,grinsztajn2022tree, marton2024grande}.

Traditional tree models typically rely on greedy algorithms or gradient-based optimization methods to learn a tree \cite{marton2024gradtree,zharmagambetov2021non}. In contrast, this paper proposes a novel decision tree model called the Learning Hyperplane Tree (LHT), which does not rely on greedy splits or gradient-based optimization. Instead, it uses hyperplanes for data partitioning. LHT employs a multi-stage splitting strategy that progressively refines data partitioning to improve the distinction between target and non-target class samples. Although the splits at each stage may not achieve perfect separation, LHT gradually enhances sample separation by introducing new hyperplanes with each branching, offering a more structured and transparent partitioning process.
\textcolor{black}{LHT adopts a piecewise linear structure, ensuring high interpretability and enabling clear tracking of each feature's contribution to the decision-making process.}
Moreover, LHT demonstrates superior performance. Thanks to its piecewise linear structure, it boasts extremely fast inference speed, and its test accuracy surpasses existing SOTA methods on multiple public datasets.

\section{Learning Hyperplane Tree}
\begin{figure*}[ht]
\vskip 0.1in
\begin{center}
\centerline{\includegraphics[width=0.75\textwidth]{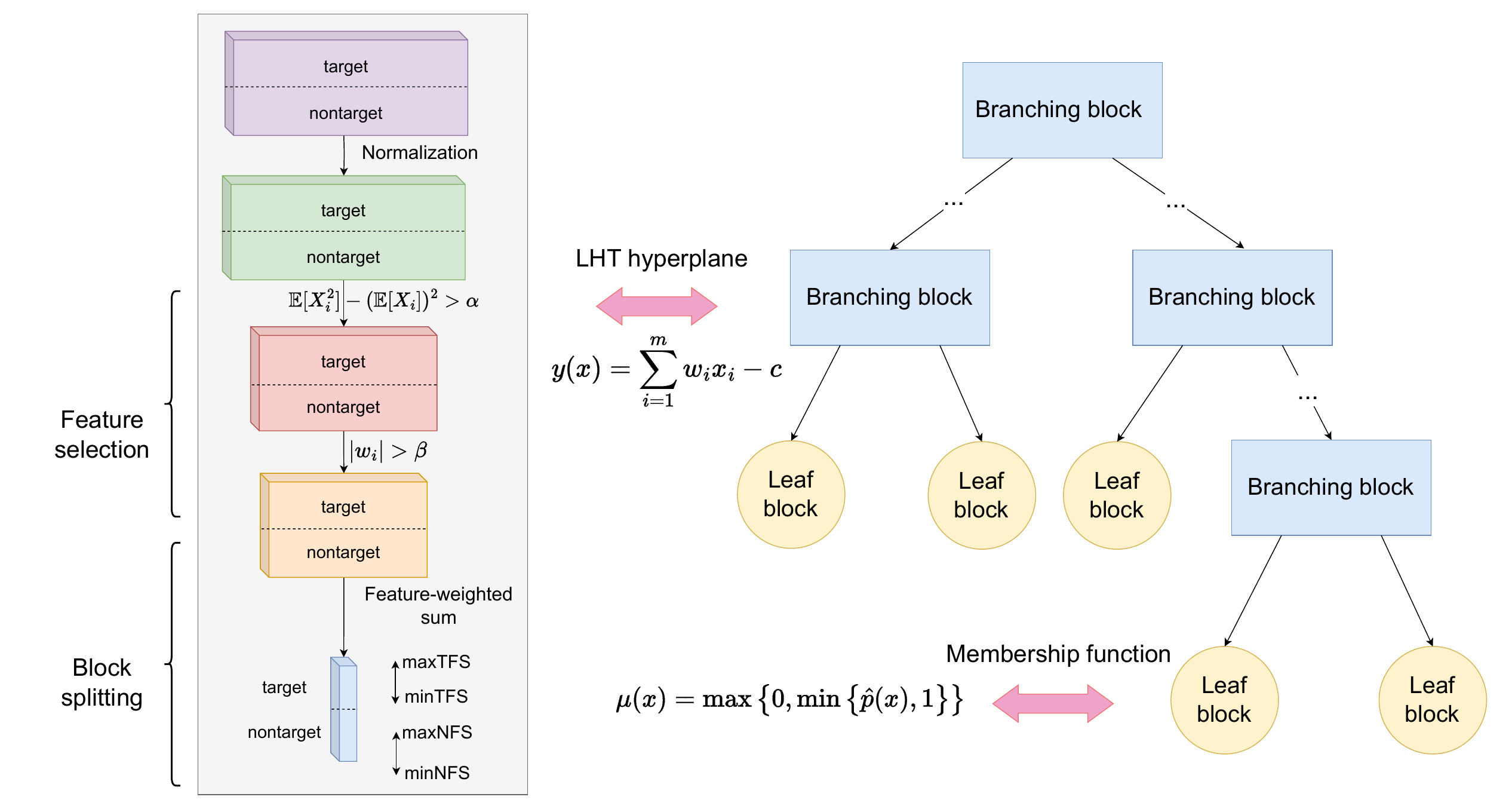}}
\caption{
The structure of LHT is illustrated. LHT consists of two types of blocks: a branching block, which employs hyperplanes for sample partitioning, and a leaf block, where least-squares fitted membership functions are used for classifying test samples.}
\label{LHT}
\end{center}
\vskip -0.3in
\end{figure*}

In this paper, each LHT addresses a binary classification problem. For a classification task, regardless of the number of classes, an LHT can be constructed for each class. Each LHT treats the samples of the corresponding class as target samples and the remaining samples as non-target samples, thereby solving a binary classification problem for that class. Different LHTs are designed to handle distinct binary classification tasks. By aggregating the results of all the LHTs, the overall classification problem can be effectively solved. As depicted in Figure \ref{LHT}, LHT originates from a root block and branches outward. The blocks within the LHT can be categorized into two types: branching blocks, which have completed their branching, and leaf blocks, which are the terminal blocks. Unlike conventional tree models, each block in the LHT typically defines two types of functions. One of these functions is a hyperplane that partitions the sample data, while the other is a membership function derived via least squares fitting and fuzzy logic. 

\subsection{LHT hyperplane}

The LHT hyperplane tend to separate target samples from non-target samples using a linear decision boundary. Each branching block in the LHT has a hyperplane that partitions the data in the block into two subsets. These subsets are then assigned to the two subblocks resulting from the split. 
Typically, both sides of the split contain a mixture of both classes. However, the hyperplane can be chosen such that one side contains only target samples or only non-target samples.
The subblock that contains pure samples is labeled as a leaf block and will not be split further. The subblock with mixed samples will continue to be split.
The data partitioning of a branching block consists of two steps: feature selection and block splitting. \textcolor{black}{\textcolor{black}{Section} \ref{feature1} provides the details of feature selection, while \textcolor{black}{Section} \ref{block2} describes the branching block splitting process in detail.}

\subsubsection{Feature Selection}
\label{feature1}
For a branching block, suppose the block has $n$ samples, each with $m$ features. $\mathbb{E}[X_i]$ represents the expected value of the $i$-th feature over the $n$ samples. Features with low usefulness can be filtered out by calculating the variance, given by the condition
\begin{equation}
\mathbb{E}[X_i^2] - (\mathbb{E}[X_i])^2 > \alpha, \:i={1,2,\cdots,m}.
\end{equation}
where $\alpha$ is a nonnegative constant.

For the block, $\mathbb{E}[X_i]^{\text{t}}$ denotes the expected value of the $i$-th feature for the target samples, and $\mathbb{E}[X_i]^{\text{nt}}$ represents the expected value of the $i$-th feature for the non-target samples. The quantity $\text{SD}_i$ is defined as the difference between the expected feature values of target and non-target samples:
\begin{equation}
	\text{SD}_i=\mathbb{E}[X_i^{\text{t}}]-\mathbb{E}[X_i^{\text{nt}}], \:i={1,2,\cdots,m}.
	\label{SD}
\end{equation}
$\text{SD}_i$ measures the feature difference between target and non-target samples for the $i$-th feature and can be either positive or negative. A smaller absolute value of $\text{SD}_i$ indicates that the target and non-target samples are less distinguishable based on the $i$-th feature, suggesting that \textcolor{black}{the two sample types may not be significantly differentiated through this feature.} Conversely, a larger absolute value of $\text{SD}_i$ reflects a more pronounced feature difference, highlighting that this feature is more important in distinguishing between the target and non-target samples.

Based on $\text{SD}_i$, the quantity $\max \text{SD}$ is defined as:
\begin{equation}
	\max \text{SD} = \max\{\text{SD}_1, \text{SD}_2, \cdots, \text{SD}_m\},
	\label{maxSD}
\end{equation}
Similarly, $\min \text{SD}$ is defined as:
\begin{equation}
	\min \text{SD} = \min\{\text{SD}_1, \text{SD}_2, \cdots, \text{SD}_m\}.
	\label{minSD}
\end{equation}
The quantity $\overline{\text{SD}}$ represents the maximum absolute value of the difference between $EX_i^{\text{t}}$ and $EX_i^{\text{nt}}$ across all features, indicating the features with the most significant differences. It is defined as:
\begin{equation}
	\overline{\text{SD}} =\max \{|\min \text{SD}|, |\max \text{SD}|\}.
	\label{absSD}
\end{equation}

The weight coefficient of feature $i$, denoted as $w_i$, is defined as:
\begin{equation}
	w_i=
		\frac{\text{SD}_i}{\overline{\text{SD}}}.
	\label{feature_weight}
\end{equation}
Note that $-1\leq w_i\leq1$.
Since $w_i$ reflects the distinguishing capability of the corresponding feature between the two classes, features with higher distinguishing ability can be further selected using the criterion $|w_i|>\beta$, where $0\leq\beta<1$. When the value of $\beta$ is large, fewer features are selected, but the chosen features tend to have higher discriminative power. On the other hand, when $\beta$ is small, more features are selected.

\subsubsection{Block Splitting}
\label{block2}
For a input data vector $x\in \mathbb{R}^m$, we define the feature-weighted sum as:
\begin{equation}
	\text{FS}(x)=\sum\limits_{i=1}^mw_ix_i.
	\label{feature_sum}
\end{equation}
The hyperplane associated with the block is expressed as:
\begin{equation}
		y(x)=\text{FS}(x)-c,
	\label{hyperplane}
\end{equation}
\textcolor{black}{where $c$ is a constant, the selection of which will be described later.}
In the block, the sample data is partitioned into the two subblocks by checking whether $y(x) < 0$ holds.

For the $n$ samples in the block, the $i$-th feature of the $j$-th sample is denoted as $x_{ij}$. The feature-weighted sum of the $j$-th sample is
\begin{equation}
	\text{FS}_j(x)=\sum\limits_{i=1}^mw_ix_{ij}.
\end{equation}
A set of $n$ feature-weighted sums can be obtained, denoted as $\mathcal{FS}=\{\text{FS}_1,\text{FS}_2,\cdots,\text{FS}_n\}$. This set consists of the feature-weighted sums for both the target and non-target classes, i.e., $\mathcal{FS}=\mathcal{FS}^\text{t}\cup\mathcal{FS}^{\text{nt}}$. The maximum and minimum feature-weighted sums among the target samples are denoted as $\max \text{TFS}=\max\{\mathcal{FS}^\text{t}\}$ and $\min \text{TFS}=\min\{\mathcal{FS}^\text{t}\}$, respectively. Similarly, the maximum and minimum feature-weighted sums among the non-target samples are denoted as $\max \text{NFS}=\max\{\mathcal{FS}^\text{nt}\}$ and $\min \text{NFS}=\min\{\mathcal{FS}^\text{nt}\}$. These \textcolor{black}{four values as well as $e=(\min \mbox{NFS}+\max \mbox{NFS}+\min \mbox{TFS}+\max \mbox{TFS})/4$} serve as candidate constants for $c$ in the hyperplane.

\textcolor{black}{When splitting a block, \textcolor{black}{the one from the five candidate values} that most effectively separates the target samples from the non-target samples is selected as the \textcolor{black}{constant $c$ in the hyperplane (\ref{hyperplane})}. Specifically, among the two subblocks created by the split, if a leaf block contains samples of only one class, the number of samples in this leaf block should be maximized. }The counting is first performed as follows:
\begin{equation}
	\begin{aligned}
		N_1 &= \left| \{ j \mid \text{FS}_j \in \mathcal{FS}^\text{t} , \text{FS}_j < \min \text{NFS} \} \right|, \\
		N_2 &= \left| \{ j \mid \text{FS}_j \in \mathcal{FS}^\text{t} ,  \text{FS}_j > \max \text{NFS} \} \right|, \\
		N_3 &= \left| \{ j \mid \text{FS}_j \in \mathcal{FS}^\text{nt} , \text{FS}_j < \min \text{TFS} \} \right|, \\
		N_4 &= \left| \{ j \mid \text{FS}_j \in \mathcal{FS}^\text{nt} , \text{FS}_j > \max \text{TFS} \} \right|.
	\end{aligned}
\end{equation}
Here, $|\cdot|$ denotes the number of elements in a set. The expression for $c$ is given by:
\begin{equation}
	c=\begin{cases}
		\min \text{NFS},\quad N_1=N_{\max}\:\text{and}\: N_{\max}\geq\gamma\\
		\max \text{NFS},\quad N_2=N_{\max}\:\text{and}\: N_{\max}\geq\gamma\\
		\min \text{TFS},\quad N_3=N_{\max}\:\text{and}\: N_{\max}\geq\gamma\\
		\max \text{TFS},\quad N_4=N_{\max}\:\text{and}\: N_{\max}\geq\gamma\\
        e,\quad N_{\max}<\gamma
	\end{cases}
	\label{feature_weight1}
\end{equation}
where $N_{\max}=\max\{N_1, N_2, N_3, N_4\}$.
When the maximum number of samples that can be completely separated for a certain class is smaller than the threshold $\gamma$, the value of $c$ is set to the average of the four candidate constant terms of the hyperplane, denoted as $e$. \textcolor{black}{The parameter $e$ is set to ensure that both subblocks resulting from each split contain a certain number of samples. It is worth noting that if $c$ is set to $e$, both subblocks may be branching blocks; otherwise, at least one of the subblocks will be a leaf block.}

\begin{figure}[ht]
\vskip 0.2in
\begin{center}
\centerline{\includegraphics[width=0.45\textwidth]{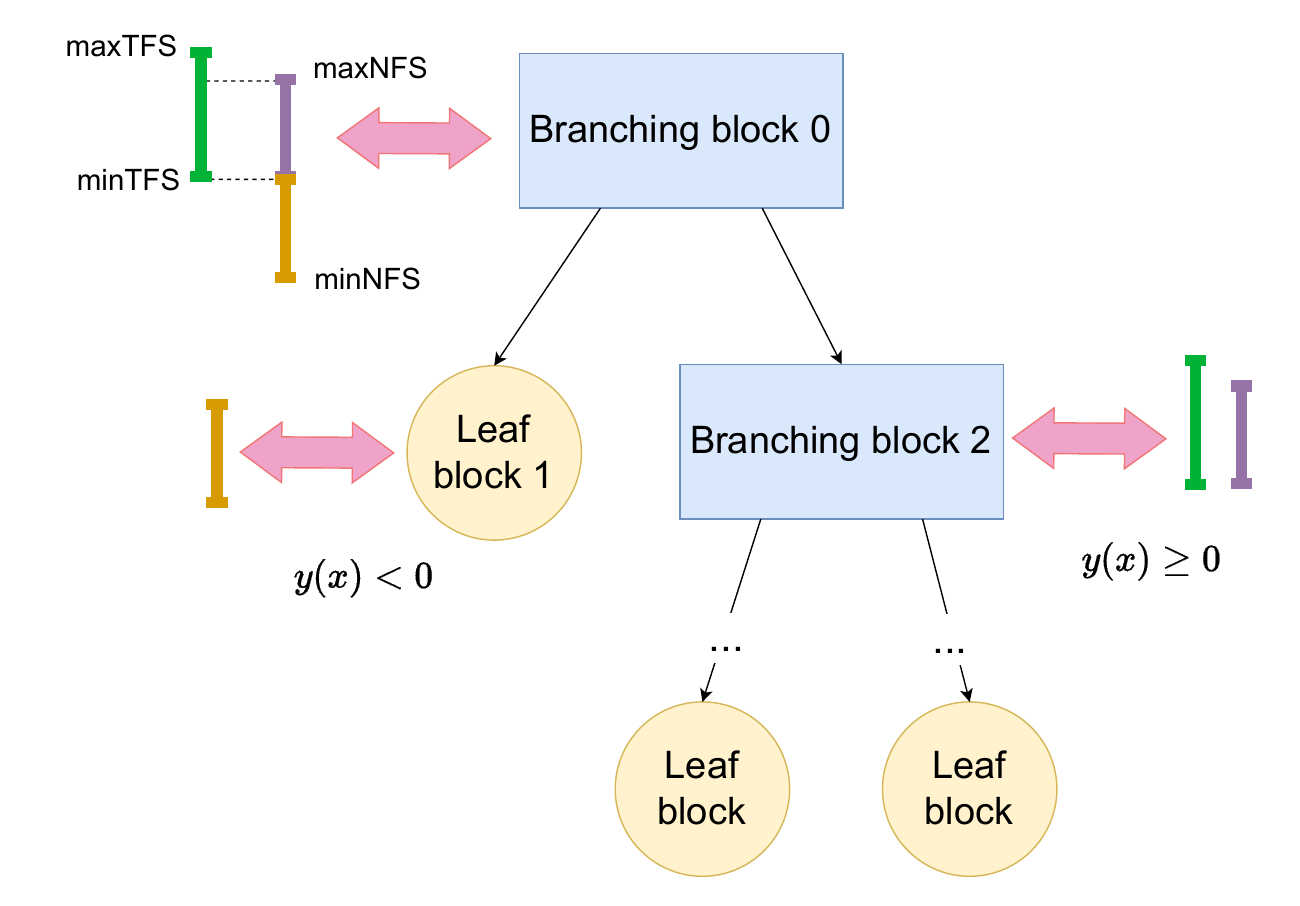}}
\caption{The case is illustrated when $N_3 = N_{\max}$ and $N_{max}\geq\gamma$, where $c = \min \text{TFS}$. Samples with a feature-weighted sum smaller than $c$ are assigned to the left subblock 1, and the remaining samples are assigned to the right subblock 2. Since all samples in left subblock 1 are non-target class samples, it is marked as a leaf block. Right subblock 2 still contains mixed samples, and the allocation process continues based on the data within the block until all samples are properly classified.}
\label{LHT_sample}
\end{center}

\end{figure}

Figure \ref{LHT_sample} presents an example of sample allocation. Branching block 0 contains both target class and non-target class samples. The feature-weighted sum for each sample is computed to obtain the feature set $\mathcal{FS}$, as well as four specific feature-weighted sums: $\min \text{TFS}$, $\max \text{TFS}$, $\min \text{NFS}$, and $\max \text{NFS}$. 
\textcolor{black}{
Next, under the condition that $N_{\max}\geq\gamma$, select the one from $\min \text{TFS}$, $\max \text{TFS}$, $\min \text{NFS}$, and $\max \text{NFS}$ that results in the largest number of samples in the leaf blocks after splitting, and use it as the current branching block hyperplane parameter $c$. Then, check the value of $y(x)$, assigning samples with $y(x)<0$ to the left subblock and the remaining samples to the right subblock. If a subblock contains samples from only one class, it is marked as a leaf block.
It is important to note that since the samples in block 2 are the remaining samples of block 0 after excluding those in block 1, $\mathbb{E}[X_i]^{\text{t}}$, $\mathbb{E}[X_i]^{\text{nt}}$, $\text{SD}_i$ and $w_i$ in block 2 change.} Consequently, the feature-weighted sum for the same sample may vary on different blocks. The LHT block splitting algorithm is given in Algorithm \ref{alg1}.

The confidence level is defined as:
\begin{equation}
		d(x)=\begin{cases}
			|y(x)|,\quad|y(x)|<1\\
			1, \quad |y(x)|\geq 1
		\end{cases}
	\label{zhixindu}
\end{equation}
$d(x)$ measures the distance of data point $x$ from the decision boundary.
The algorithm for finding the corresponding leaf block of the input data is given in Algorithm \ref{alg2}.

\begin{algorithm}[tb]
    \caption{LHT Block Splitting Algorithm - Training} 
    \label{alg1} 
    \begin{algorithmic}[1]
        \STATE {\bfseries Input:} Data within the block.
        \STATE Retrieve the data from the block.
        \STATE Normalize the features.
        \STATE Filter out features with small fluctuations based on the condition $\mathbb{E}[X_i^2] - (\mathbb{E}[X_i])^2 > \alpha$, identifying active features
        \STATE Compute the difference between target and non-target classes for the active features using (\ref{SD}).
        \STATE Calculate the weights of the active features using (\ref{feature_weight}).
        \STATE Remove active features with smaller differences based on the criterion $|w_i|>\beta$.
        \STATE Compute feature-weighted sums for remaining features using (\ref{feature_sum}).
        \STATE Select an appropriate constant $c$ from the four candidates and construct the hyperplane for the block as defined in (\ref{hyperplane}).
        \FOR{each sample $x$ in the block}
            \IF{$y(x) < 0$}
                \STATE Assign the data to the left subblock. If the assigned data is insufficient or belongs entirely to a single class, mark it as a leaf block.
            \ELSE
                \STATE Assign the data to the right subblock. If the assigned data is insufficient or belongs entirely to a single class, mark it as a leaf block.
            \ENDIF
        \ENDFOR
    \STATE {\bfseries Output:} \textcolor{black}{Two subblocks with assigned data.}
    \end{algorithmic}
\end{algorithm}

\begin{algorithm}[tb]
    \caption{Data Search Block Algorithm - Testing} 
    \label{alg2} 
    \begin{algorithmic}[1]
        \STATE {\bfseries Input:} Data $x$ at the root block.
        \WHILE{the block is not a leaf block}
            \STATE Compute $y(x)$ for the block.
            \IF{$y(x) < 0$}
                \STATE Move to the left subblock and calculate $d(x)$ using (\ref{zhixindu}).
                \IF{the left subblock is a leaf block}
                    \STATE break;
                \ENDIF
            \ELSE
                \STATE Move to the right subblock and calculate $d(x)$ using (\ref{zhixindu}).
                \IF{the right subblock is a leaf block}
                    \STATE break;
                \ENDIF
            \ENDIF
        \ENDWHILE
        \STATE The final leaf block reached is the block to which the input data $x$ is assigned, with $d(x)$ representing the confidence.
        \STATE {\bfseries Output:} \textcolor{black}{Leaf block for data $x$.}
    \end{algorithmic}
\end{algorithm}

\subsection{Membership Function}

\textcolor{black}{The membership function of an LHT is a piecewise linear function derived through least squares fitting and fuzzy logic.} Each block in the LHT can compute its corresponding membership function. The classification task can be completed using the membership function of the leaf block corresponding to the input data.

\textcolor{black}{For a labeled classification problem, the labels corresponding to the $n$ samples are denoted as ${p_1, p_2, \cdots, p_n}$, where $p_i \in \{0, 1\}, i \in {1, 2, \cdots, n}$. Here, $p = 1$ indicates the target class label, while $p = 0$ represents the non-target class label. In the case of a multi-class classification problem, multiple LHTs are constructed, each corresponding to a specific class. For each LHT, the class it corresponds to is treated as the target class, while all other classes are considered non-target classes.}

\textcolor{black}{Let $P=[p_1, p_2,\cdots,p_n]^\top$ represent the label vector, and $W=[a_1,\cdots,a_m,b]^\top$ denote the weight coefficients of the linear function. The feature matrix is given by:
$$
	X
	=
	\left[\begin{array}{llll}
		x_{11}&\cdots & x_{1m} &1 \\
		x_{21}&\cdots & x_{2m} &1\\
		\vdots&\ddots &\vdots &\vdots\\
		x_{n1}&\cdots & x_{nm}&1
	\end{array}\right]
$$
 The probability function for the samples can be approximated using a linear function, expressed as:
\begin{equation*}
	  X W \rightarrow P.
\end{equation*}
 The optimal weights vector $W^*$ can be obtained by solving the following optimization problem:
\begin{equation}
	\min \:\: (P-XW)^\top(P-XW).
\end{equation}
}

The derivation details are provided in Appendix. We get,
$$
a_i^*=\frac{\mathbb{E}[X_iP]-\mathbb{E}[X_i]\mathbb{E}[P]}{\mathbb{E}[X_i^2]-(\mathbb{E}[X_i])^2}, i={1,2,\cdots,m}
$$
$$
b^*=\mathbb{E}[P]-\sum\limits_{i=1}^m a_i^*\mathbb{E}[X_i],
$$
\textcolor{black}{in which $\mathbb{E}[X_i]=(x_{1i}+\cdots+x_{ni})/n$, $\mathbb{E}[X_i^2]=(x_{1i}^2+\cdots+x_{ni}^2)/n$, $\mathbb{E}[P]=(p_{1}+\cdots+p_{n})/n$, $\mathbb{E}[X_iP]=(x_{1i}p_{1}+\cdots+x_{ni}p_{n})/n$, $i,j={1,2,\cdots,m}$.}
Hence, we have:
\begin{equation}
  \hat{p}(x)=\sum\limits_{i=1}^m a_i^*(x_i-\mathbb{E}[X_i]) +\mathbb{E}[P].
\end{equation}
\textcolor{black}{To output a probability value between 0 and 1, the membership function is defined using min-max operations (fuzzy logic) as follows:}
\begin{equation}
\mu(x)=\max\left\{0,\min\left\{\hat{p}(x),1\right\}\right\}.
\end{equation}

The confidence $d(x)$ measures the distance of the data 
$x$ from the boundary and can be used as a correction term for the membership function. The corrected membership function is given by:
\begin{equation}
	d(x)\cdot\mu(x).
\end{equation}

\section{LH forest}

When the performance of a single LHT is limited, an LH forest composed of multiple LHTs can be constructed to better extract data features and achieve improved classification results.
For scenarios with a large number of samples but relatively few features, a method similar to random forest construction can be adopted, where multiple LHTs are generated by randomly sampling subsets of data to form the LH forest. In contrast, for cases with fewer samples but a larger number of features, different feature subsets can be selected to build multiple LHTs for classification tasks.
It is worth noting that the feature weights $w_i$ of LHTs reflect the distinguishing capability of individual features in classification, providing valuable guidance for effective feature selection.
The features can be selected by setting $\beta = \frac{i}{t}$, where $i \in \{0, 1, \cdots, t-1\}$, ensuring diversity in the features chosen for each tree, while consistently selecting those features with higher discriminative ability.

If there are $t$ LHTs used for the classification of data $x$, the final corrected membership function for the target class is:
\begin{equation}
	\frac{1}{t}\sum\limits_{i=1}^t d_i(x)\cdot\mu_i(x),
    \label{wolf}
\end{equation}
where the set of linear functions described in (\ref{wolf}) is referred to as Weighted Overlapping Linear Functions (WOLF). 
\textcolor{black}{For a single LHT, when all blocks are taken together, their linear functions form a discontinuous membership function over the whole measurement space. By overlapping and weighting the outputs of several LHTs, the discontinuities can be removed and the accuracy increased.}

\section{Interpretability Mechanisms}

LHT distinguishes target class samples from non-target class samples by branching multiple times through different hyperplanes. The membership functions are represented by piecewise linear functions. Moreover, the sample features in the LHT are only subjected to simple normalization without any additional feature extraction. As a result, the overall structure of LHT remains piecewise linear. This inherent structural transparency makes LHT fully interpretable.

In the LHT, each branching block corresponds to a hyperplane, with the coefficients representing the feature weights $w_i$. The absolute value of $w_i$ indicates the significance of the corresponding feature in distinguishing between target and non-target class samples. Therefore, each branching block generated by the LHT reveals the contribution of each feature to the classification of the current sample, as reflected by the magnitude of $|w_i|$.

To illustrate the contribution of each feature in the LHT branching, we use the wine dataset as an example. The wine dataset consists of 178 samples and 13 features, and it is a multi-class classification problem with three classes. 80\% of the data is used for training, and 20\% is used for testing. For each class, a separate LHT is trained, resulting in a total of three LHTs for classification. The results of the LHT classification, under the condition of fixed $\alpha=0$ and varying $\beta$, are shown in Table \ref{sample-table1}.
The value of $\beta$ can influence the growth of LHT in different ways. For the wine dataset, selecting fewer but more discriminative features (i.e., using a larger $\beta$) does not necessarily lead to a decrease in test accuracy. In six tests, the test accuracy was 100\% for $\beta = 0, 0.25, 0.5$, while for $\beta = 0.7$, the accuracy dropped to 97.30\%. However, when $\beta = 0.8$, the test accuracy increased again to 100\%.

\begin{table}[t]
	\caption{Classification accuracy of the LHT on the wine dataset for varying $\beta$.}
	\label{sample-table1}
	\centering
	\begin{center}
		\begin{small}
			\begin{sc}
				\begin{tabular}{>{\centering\arraybackslash}c>{\centering\arraybackslash}p{2.1cm}>{\centering\arraybackslash}p{2.1cm}}
					\toprule
					\multirow{2}{*}{$\beta$} & Training Accuracy(\%)& Testing Accuracy(\%)\\
					\midrule
				    0 & 100 &100\\
				    0.25 & 100& 100\\
                        0.5 & 99.29& 100\\
				    0.7 & 99.29& 97.30\\
				    0.8 & 99.29& 100\\
					\bottomrule
				\end{tabular}
			\end{sc}
		\end{small}
	\end{center}
	\vskip -0.3in
\end{table}

\begin{figure*}[ht]
\vskip 0.1in
\begin{center}
\centerline{\includegraphics[width=0.95\textwidth]{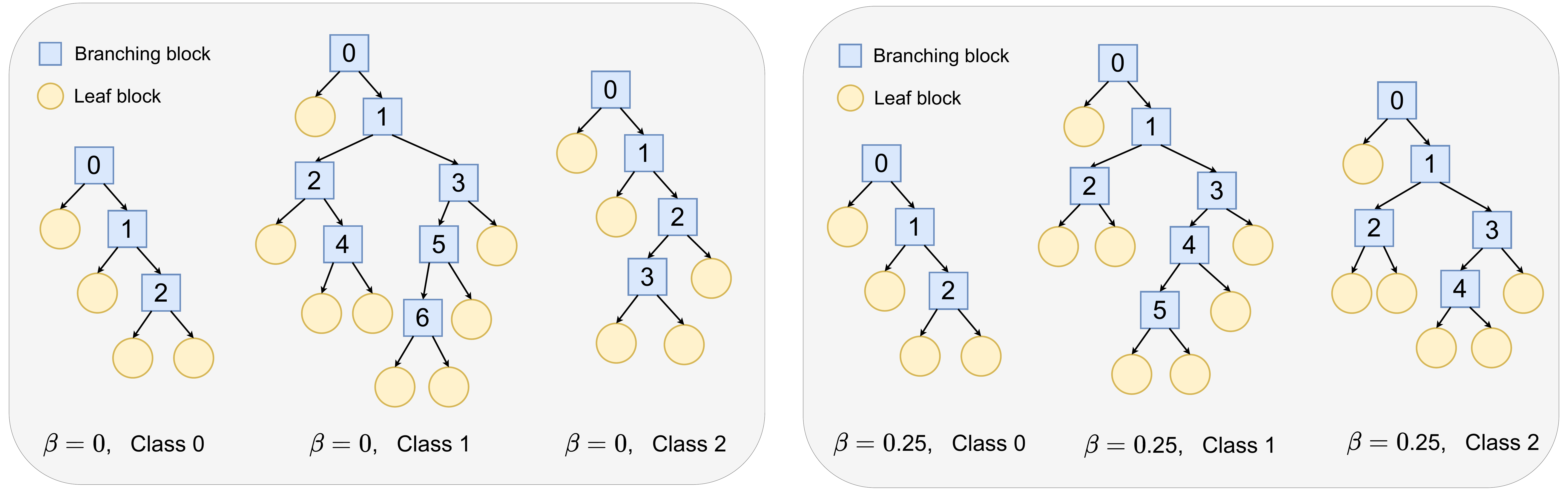}}
\caption{The LHT structures of the three classes in the wine dataset are shown, with the left side corresponding to the case where $\beta=0$, and the right side to the case where $\beta=0.25$.}
\label{LHT_s0}
\end{center}
\vskip -0.2in
\end{figure*}

The LHT structures of the three classes in the wine dataset are shown in Figure \ref{LHT_s0}, with the left side corresponding to the case where $\beta=0$, and the right side to the case where $\beta=0.25$.
For class 0, the LHT structures are the same for both $\beta=0$ and $\beta=0.25$. For class 1, the number of blocks decreases when $\beta=0.25$, while for class 2, the number of blocks increases when $\beta=0.25$.

\begin{figure*}[htbp]
\vskip -0.1in
	\centering    
	\subfigure[$\beta=0$] 
	{
		\begin{minipage}[t]{0.36\linewidth}
			\centering          
			\includegraphics[width=1\textwidth]{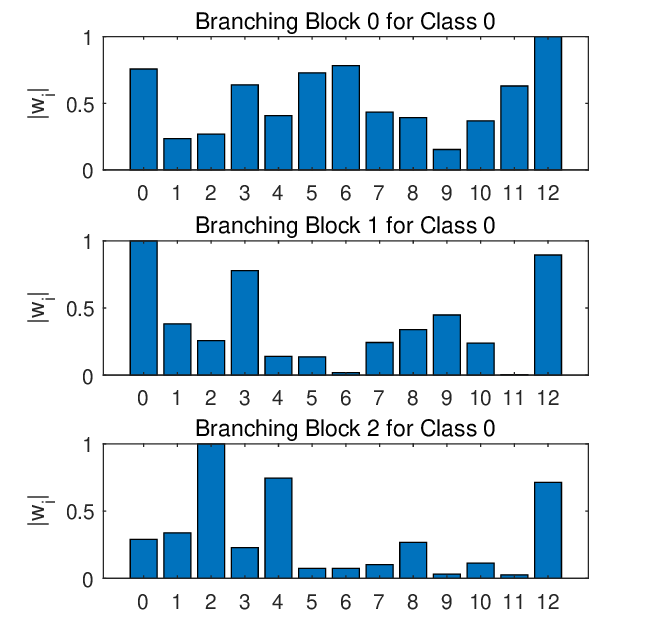}   
      
		\end{minipage}
	}
	\subfigure[$\beta=0.25$] 
	{
		\begin{minipage}[t]{0.36\linewidth}
			\centering      
			\includegraphics[width=1\textwidth]{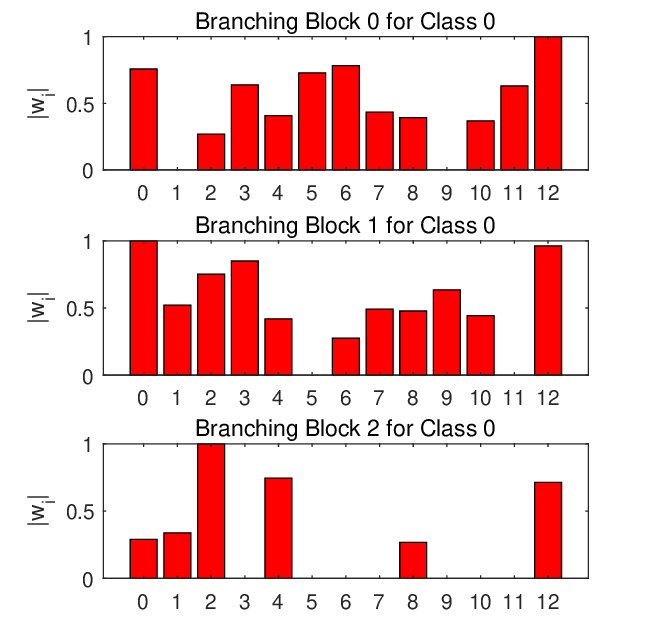}   
          
		\end{minipage}
	}
    \vskip -0.1in
	\caption{The LHT feature weight visualization for class 0 of the wine dataset is shown, with the left side corresponding to the case of $\beta=0$ and the right side to the case of $\beta=0.25$.} 
	\label{LHT_f}  
\end{figure*}

For the LHT corresponding to class 0 in the wine dataset, Figure \ref{LHT_f} demonstrates the contribution of each feature to the classification of the branching blocks. In the split of branching block 0, feature 12 is the most influential. For branching block 1, feature 0 plays the most significant role, whereas feature 2 dominates in the split of branching block 2.
The situations for class 1 and class 2 are similar, as detailed in Appendix.
Due to the transparency of the LHT structure, the contribution of each feature at every branching step is clearly visible, which is why we refer to LHT as fully interpretable.

\section{Experiments}
\begin{table*}[t]
	\caption{Classification accuracies on small-sample datasets.}
	\label{sample-tables}
	\centering
	\begin{center}
		\begin{small}
			\begin{sc}
				\begin{tabular}{>{\centering\arraybackslash}p{1.8cm}>{\centering\arraybackslash}c>{\centering\arraybackslash}p{0.5cm}>{\centering\arraybackslash}p{2cm}>{\centering\arraybackslash}p{2.3cm}>{\centering\arraybackslash}p{2.3cm}>{\centering\arraybackslash}p{2cm}>{\centering\arraybackslash}p{2cm}}
					\toprule
					\multirow{2}{*}{Data \par set} &\multirow{2}{*}{$n$}&\multirow{2}{*}{$m$} & \multirow{2}{*}{Method} & Training accuracy(\%)$\uparrow$ & Testing accuracy(\%) $\uparrow$& Training time($ms$)$\downarrow$ & Testing time($ms$)$\downarrow$ \\
					\midrule
					    \multirow{6}{*}{wine}& \multirow{6}{*}{178}&\multirow{6}{*}{13}& CART &100&94.40 &\bf{1.0}&0.027  \\
				    &&&  RF&  100&94.40  &47&2.975\\
                       &&&  XGboost&  100&97.22  &27&0.364\\
                       &&&  catboost&  100&100 &786&0.396\\
                       &&&  lightGBM&  100&100 &73&0.584 \\
                       &&& LHT&\bf{100}& \bf{100} &1.4 &\bf{0.011}\\
					   \midrule
                    \multirow{6}{*}{seeds} &\multirow{6}{*}{ 210 }&\multirow{6}{*}{7 }
                          &  CART &   95.83  & 88.10 & \bf{1.1} & 0.035\\
				    &&&  RF&  96.43& 88.10 & 6 &0.429\\
                       &&&  XGboost&  98.81& 95.24 &9.6&0.285\\
                       &&&  catboost&  98.81&92.86 &301 &0.354\\
                       &&&  lightGBM&  100&90.48& 111 &0.441 \\
                       &&&  LHT& \bf{100} &\bf{97.62} &2.3&  \bf{0.013}  \\
					\midrule
					\multirow{6}{*}{WDBC} &\multirow{6}{*}{569}&\multirow{6}{*}{30}& CART 
                        &99.78&93.86 & \bf{5.1} &0.052  \\
				    &&&  RF&  99.56&96.49  & 70&3.069\\
                       &&&  XGboost&  100&  95.61&71&0.285\\
                       &&&  catboost&  99.56&97.37  &2348&0.544\\
                       &&&  lightGBM&  100&97.37  &73&0.442\\
                       &&&  LHT&  \bf{100}&   \bf{100} &11.4&\bf{0.023}\\
                         \midrule
                      \multirow{6}{*}{\parbox{2cm}{\centering banknote}} &\multirow{6}{*}{ 1372 }&\multirow{6}{*}{3 }
                          &  CART &98.72&     96.73 &\bf{2.1}&0.045 \\
				    &&&  RF& 99.82 & 98.91 &37&1.534\\
                       &&&  XGboost&  100&  99.27&21&0.281\\
                       &&&  catboost&  100&99.64 &462&0.354\\
                       &&&  lightGBM&  100&99.64  &76&0.515\\
                       &&&  LHT&  \bf{100}&     \bf{100}&4.7&\bf{0.016}\\
					\bottomrule
				\end{tabular}
			\end{sc}
		\end{small}
	\end{center}
	\vskip -0.1in
\end{table*}

\begin{table*}[t]
	\caption{Classification accuracies on medium-scale and large-scale datasets.}
	\label{sample-tablel}
	\centering
	\begin{center}
		\begin{small}
			\begin{sc}
				\begin{tabular}{>{\centering\arraybackslash}p{1.8cm}>{\centering\arraybackslash}c>{\centering\arraybackslash}p{0.5cm}>{\centering\arraybackslash}p{2cm}>{\centering\arraybackslash}p{2.3cm}>{\centering\arraybackslash}p{2.3cm}>{\centering\arraybackslash}p{2cm}>{\centering\arraybackslash}p{2cm}}
					\toprule
					\multirow{2}{*}{Data \par set} &\multirow{2}{*}{$n$}&\multirow{2}{*}{$m$} & \multirow{2}{*}{Method} & Training accuracy(\%) $\uparrow$& Testing accuracy(\%) $\uparrow$& Training time($ms$)$\downarrow$& Testing time($ms$) $\downarrow$\\
                            \midrule
                                        \multirow{6}{*}{Rice} &\multirow{6}{*}{3810  }&\multirow{6}{*}{7 }
                          &  CART &   94.69  &  91.47&\bf{10}&0.047 \\
				    &&&  RF&  95.08& 92.52 &68&1.236\\
                       &&&  XGboost&  96.59& 92.26 &48&0.307\\
                       &&&  catboost&  94.26& \bf{92.78} &669&0.377\\
                       &&&  lightGBM&  95.83&92.39 &91&0.475 \\
                       &&&  LHT& \bf{97.05} &  92.26 &49&\bf{0.016}  \\
					\midrule
                    \multirow{6}{*}{Spambase} &\multirow{6}{*}{4601}&\multirow{6}{*}{57}
                            & CART &92.91&   91.10  &\bf{16}&0.046   \\
				    &&&  RF&  93.07& 91.64 &261&5.529\\
                       &&&  XGboost&  99.08& 94.90 &189&0.309\\
                       &&&  catboost&  98.97& 94.90 &2090&0.602\\
                       &&&  lightGBM&  98.26&95.11  &117&0.507\\
                       &&&  LHT&  \bf{99.13}&    \bf{95.44} &476&\bf{0.026}\\
					\midrule
                    \multirow{6}{*}{EEG} &\multirow{6}{*}{ 14980 }&\multirow{6}{*}{14 }
                          &  CART &    97.08 & 83.01  &\bf{86}&0.044\\
				    &&&  RF& 98.65 & 90.65 &1606&5.633\\
                      &&&  XGboost&  \bf{100}&  94.06&427&0.334\\
                       &&&  catboost&  99.76& 94.46 &7444&0.428\\
                       &&&  lightGBM&  99.95&  93.49&114&0.519\\
                       &&&  LHT& 98.07 & \bf{94.59}  &649&\bf{0.025} \\
					\midrule
                        \multirow{6}{*}{\parbox{2cm}{\centering MAGIC\\ Gamma\\ Telescope}} &\multirow{6}{*}{ 19020 }&\multirow{6}{*}{ 10} &  CART &  85.15   & 83.94  &\bf{77}&0.047\\
				    &&&  RF&  85.62&  84.91&2012&5.494\\
                       &&&  XGboost&  98.03&  88.12&275&0.309\\
                        &&&  catboost&  96.69& \bf{88.43} &6392&0.401\\
                       &&&  lightGBM&  92.28&  \bf{88.43}&108&0.541\\
                       &&&  LHT&  \bf{99.53}&   86.72 &813&\bf{0.022}\\
					\midrule
                      \multirow{6}{*}{\parbox{2cm}{\centering SKIN-\\SEGMEN-\\TATION}} &\multirow{6}{*}{ 245057 }&\multirow{6}{*}{3 }
                          &  CART &   98.98  &98.91 &\bf{115} &0.048  \\
				    &&&  RF&  99.86&  99.83&5973&5.715\\
                       &&&  XGboost&  99.88& 99.84 &267& 0.301\\
                       &&&  catboost&  99.88& 99.88&19097&0.351\\
                       &&&  lightGBM&  \bf{99.95}&  99.89&243&0.490\\
                       &&&  LHT&  99.92&     \bf{99.91}&  843&\bf{0.017}\\
					\bottomrule
				\end{tabular}
			\end{sc}
		\end{small}
	\end{center}
	\vskip -0.1in
\end{table*}

To evaluate the performance of LHT, we test it on nine UCI public datasets and compare the results with several SOTA tree-based methods. These datasets include problems with small, medium, and large sample sizes. The results, as shown in Table \ref{sample-tables} and Table \ref{sample-tablel}, demonstrate that LHT achieves the best classification performance on seven datasets and has the fastest inference speed among all methods. In both tables, $n$ represents the number of samples, and $m$ represents the number of features.
The tests for all models were conducted using an AMD Ryzen 7 5800H processor. The comparative methods are described as follows:

\textbf{CART}: As a baseline method, we use the DecisionTreeClassifier from \texttt{scikit-learn}\footnote{\url{https://scikit-learn.org/}} \cite{pedregosa2011scikit}, with Gini impurity as the splitting criterion to construct the decision tree.

\textbf{Random Forest}: We use the RandomForestClassifier implemented in \texttt{scikit-learn}. 

\textbf{XGBoost, CatBoost, and LightGBM}: These models are implemented using their respective libraries: \texttt{XGBoost}\footnote{\url{https://xgboost.ai/}}, \texttt{CatBoost}\footnote{\url{https://catboost.ai/}}, and \texttt{LightGBM}\footnote{\url{https://lightgbm.readthedocs.io/}}.

For all models, the training and testing sets are randomly split at an 80\%/20\% ratio.
As shown in Table \ref{sample-tables}, LHT performs exceptionally well in small sample scenarios, achieving the highest test accuracy on all four small sample datasets (wine, seeds, WDBC, and banknote), with 100\% test accuracy on the wine, WDBC, and banknote datasets. Additionally, \textcolor{black}{LHT achieves the fastest testing speed among all small sample datasets and ranks second in training speed.}

On medium and large-scale datasets, LHT also demonstrates excellent performance. As shown in Table \ref{sample-tablel}, LHT achieves the fastest inference speed on all datasets. Its test accuracy is the highest among all methods on the spambase, EEG, and skin segmentation datasets.

For the banknote dataset in Table \ref{sample-tables} and the spambase, EEG, magic gamma telescope, and skin segmentation datasets in Table \ref{sample-tablel}, \textcolor{black}{multiple LHTs are built to form LH forests and improve accuracy. Since each LHT is independently constructed, to ensure fairness in the experiments, the training and testing times are taken as the longest time required among all LHTs.}

\section{Related Work}
Decision trees are commonly used tools for classification and regression in machine learning, with typical decision tree algorithms including ID3 cite{quinlan1993c4}, C4.5 \cite{quinlan1993c4}, and CART \cite{breiman1984cart}. ID3 selects features based on information gain, while C4.5 improves upon this by introducing the information gain ratio to overcome the bias of ID3 \cite{lestari2020increasing}. CART constructs decision trees by calculating Gini impurity (for classification tasks) or mean squared error (for regression tasks) and generates a binary tree structure. Although decision trees are easy to understand and implement, they are prone to overfitting and instability. To overcome these limitations, ensemble learning methods are widely applied to decision tree models, including Bagging \cite{breiman1996bagging,breiman1998arcing} and Boosting \cite{breiman1996bias}. \textcolor{black}{Random forests, a classic implementation of Bagging, mitigate overfitting by aggregating the predictions of multiple decision trees. Recent advancements in random forest research include reconstructing the training dataset \cite{pmlr-v235-ferry24a} and developing density estimation algorithms \cite{pmlr-v162-wen22c}.}
Boosting, in contrast,  trains multiple models sequentially, with each new model focusing on correcting the errors of the previous model, and ultimately combining the predictions of all models with weighted voting to improve accuracy. AdaBoost \cite{freund1997decision}, one of the earliest Boosting algorithms, focuses on iteratively adjusting sample weights to emphasize hard-to-classify samples, combining a sequence of weak classifiers into a strong classifier. Gradient boosting decision trees (GBDT) \cite{friedman2001greedy} is a variant of Boosting, optimizing the prediction error of each tree through gradient descent. XGBoost \cite{chen2016xgboost}, LightGBM \cite{ke2017lightgbm}, and CatBoost \cite{prokhorenkova2018catboost} are optimized versions of GBDT, improving both computational efficiency and accuracy.

\section{Discussion and Conclusion}
This paper presents a novel tree-based model, LHT. We evaluate the performance of LHT on nine public datasets. The experimental results show that LHT outperforms existing SOTA tree models on seven of the datasets and achieves the fastest inference speed across all datasets.

LHT exhibits remarkable performance in processing tabular data while maintaining strong interpretability. This transparency in decision-making is particularly valuable within the current context of responsible AI, where demands for fairness, transparency, and security are paramount. As such, LHT has the potential to play an increasingly significant role in addressing these challenges. Furthermore, it is worth investigating how LHT can handle homogeneous data, such as image and language, which remains an important research direction to expand its applicability in diverse AI tasks.


\bibliography{reference}
\bibliographystyle{ieeetr}

\newpage
\onecolumn
\appendix

\section{Derivation Details of Membership Functions}

\label{ap1}

For a labeled classification problem, the labels corresponding to the $n$ samples are ${p_1, p_2, \cdots, p_n}$, where $p_i \in \{0, 1\}, i \in {1, 2, \cdots, n}$. The  probability function of the samples can be approximated using a linear function, yielding:
\begin{equation}
	\hat{P} = X\cdot W,
\end{equation}
that is,
$$
	\left[\begin{array}{l}
		\hat{p}_1\\
		\hat{p}_2\\
		\vdots\\
		\hat{p}_n
	\end{array}\right]
	=
	\left[\begin{array}{llll}
		x_{11}&\cdots & x_{1m} &1 \\
		x_{21}&\cdots & x_{2m} &1\\
		\vdots&\ddots &\vdots &\vdots\\
		x_{n1}&\cdots & x_{nm}&1
	\end{array}\right]
	\cdot
	\left[\begin{array}{l}
		a_1\\
		\vdots\\
		a_m\\
		b
	\end{array}\right].
$$
Here, 
$\hat{P}$ is the approximation of 
$P$, and 
$W$ denotes the weight coefficients of the linear function. The weights $W$ can be obtained by solving the following optimization problem:
\begin{equation}
	\min \:\:\sum\limits_{i=1}^n(p_i-\hat{p}_i).
\end{equation}
The optimization problem can be rewritten in matrix form as follows:
\begin{equation}
	\min \:\: (P-XW)^\top(P-XW).
\end{equation}
Thus, the optimal $W^*$ satisfies:
\begin{equation}
	X^\top XW^*=X^\top P.
    \label{14}
\end{equation}

Equation (\ref{14}) can be expanded as:

\begin{equation*}
\left[\begin{array}{llll}
	x_{11}&\cdots & x_{1m} &1 \\
	x_{21}&\cdots & x_{2m} &1\\
	\vdots&\ddots &\vdots &\vdots\\
	x_{n1}&\cdots & x_{nm}&1
\end{array}\right]^\top
\left[\begin{array}{llll}
	x_{11}&\cdots & x_{1m} &1 \\
	x_{21}&\cdots & x_{2m} &1\\
	\vdots&\ddots &\vdots &\vdots\\
	x_{n1}&\cdots & x_{nm}&1
\end{array}\right]
\left[\begin{array}{l}
	a_1^*\\
	\vdots\\
	a_m^*\\
	b^*
\end{array}\right]
=
\left[\begin{array}{llll}
	x_{11}&\cdots & x_{1m} &1 \\
	x_{21}&\cdots & x_{2m} &1\\
	\vdots&\ddots &\vdots &\vdots\\
	x_{n1}&\cdots & x_{nm}&1
\end{array}\right]^\top
\left[\begin{array}{l}
	{p}_1\\
	{p}_2\\
	\vdots\\
	{p}_n
\end{array}\right]
\end{equation*}

\begin{equation*}
\left[\begin{array}{lllll}
n\mathbb{E}[X_1^2]&n\mathbb{E}[X_1X_2]&\cdots & n\mathbb{E}[X_1X_m] &n\mathbb{E}[X_1]\\
	n\mathbb{E}[X_1X_2]&n\mathbb{E}[X_2^2]&\cdots & n\mathbb{E}[X_2X_m] &n\mathbb{E}[X_2]\\
	\vdots&\vdots&\ddots &\vdots &\vdots\\
	n\mathbb{E}[X_1X_m]&n\mathbb{E}[X_2X_m]&\cdots & n\mathbb{E}[X_m^2]&n\mathbb{E}[X_m]\\
		n\mathbb{E}[X_1]&n\mathbb{E}[X_2]&\cdots & n\mathbb{E}[X_m]&n
\end{array}\right]
\left[\begin{array}{l}
	a_1^*\\
	\vdots\\
	a_m^*\\
	b^*
\end{array}\right]
=
\left[\begin{array}{l}
	n\mathbb{E}[X_1P]\\
	n\mathbb{E}[X_2P]\\
	\vdots\\
	n\mathbb{E}[X_mP]\\
	n\mathbb{E}[P]\\
\end{array}\right]
\end{equation*}
in which $\mathbb{E}[X_i]=(x_{1i}+\cdots+x_{ni})/n$, $\mathbb{E}[X_i^2]=(x_{1i}^2+\cdots+x_{ni}^2)/n$, $\mathbb{E}[X_iX_j]=(x_{1i}x_{1j}+\cdots+x_{ni}x_{nj})/n$, $\mathbb{E}[P]=(p_{1}+\cdots+p_{n})/n$, $\mathbb{E}[X_iP]=(x_{1i}p_{1}+\cdots+x_{ni}p_{n})/n$, $i,j={1,2,\cdots,m}$.

We get,
$$
a_i^*=\frac{\mathbb{E}[X_iP]-\mathbb{E}[X_i]\mathbb{E}[P]}{\mathbb{E}[X_i^2]-(\mathbb{E}[X_i])^2}, i={1,2,\cdots,m}
$$
$$
b^*=\mathbb{E}[P]-\sum\limits_{i=1}^m a_i^*\mathbb{E}[X_i].
$$
Hence, we have:
\begin{equation}
  \hat{p}(x)=\sum\limits_{i=1}^m a_i^*(x_i-\mathbb{E}[X_i]) +\mathbb{E}[P].
\end{equation}
To output a probability value between 0 and 1, the membership function is defined using min-max operations (fuzzy logic) as follows:
\begin{equation}
\mu(x)=\max\left\{0,\min\left\{\hat{p}(x),1\right\}\right\}.
\end{equation}


\newpage
\section{Feature Importance of the LHT Branching Block in the Wine Dataset}
\label{ap2}

\begin{figure*}[ht]
\vskip 0.2in
\begin{center}
\centerline{\includegraphics[width=0.9\textwidth]{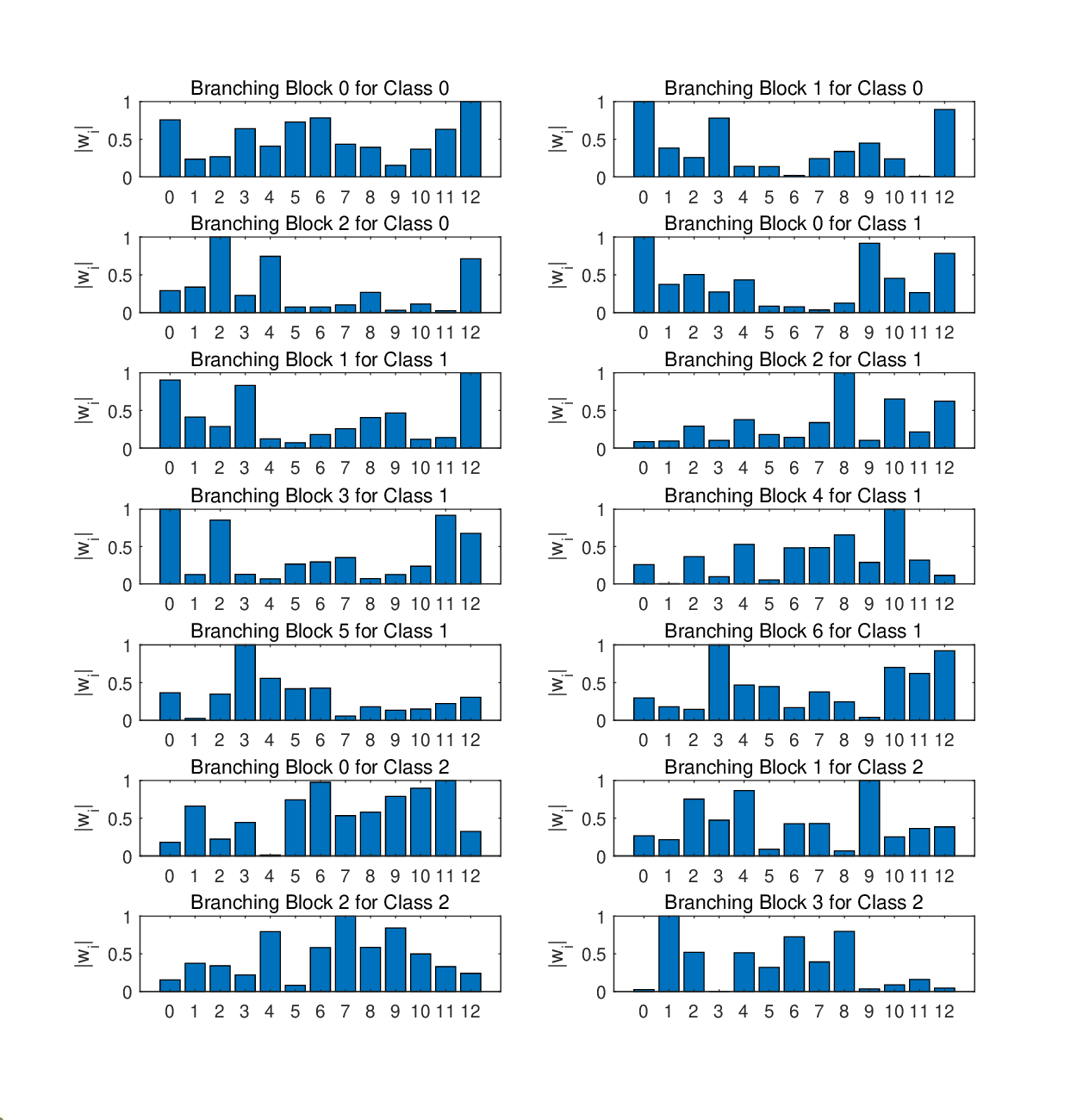}}
\caption{Visualization of the feature weights for each branching block of the three LHTs corresponding to the three classes in the wine dataset ($\beta=0$).}
\label{LHT_f14}
\end{center}
\vskip -0.2in
\end{figure*}

\begin{figure*}[ht]
\vskip 0.2in
\begin{center}
\centerline{\includegraphics[width=0.9\textwidth]{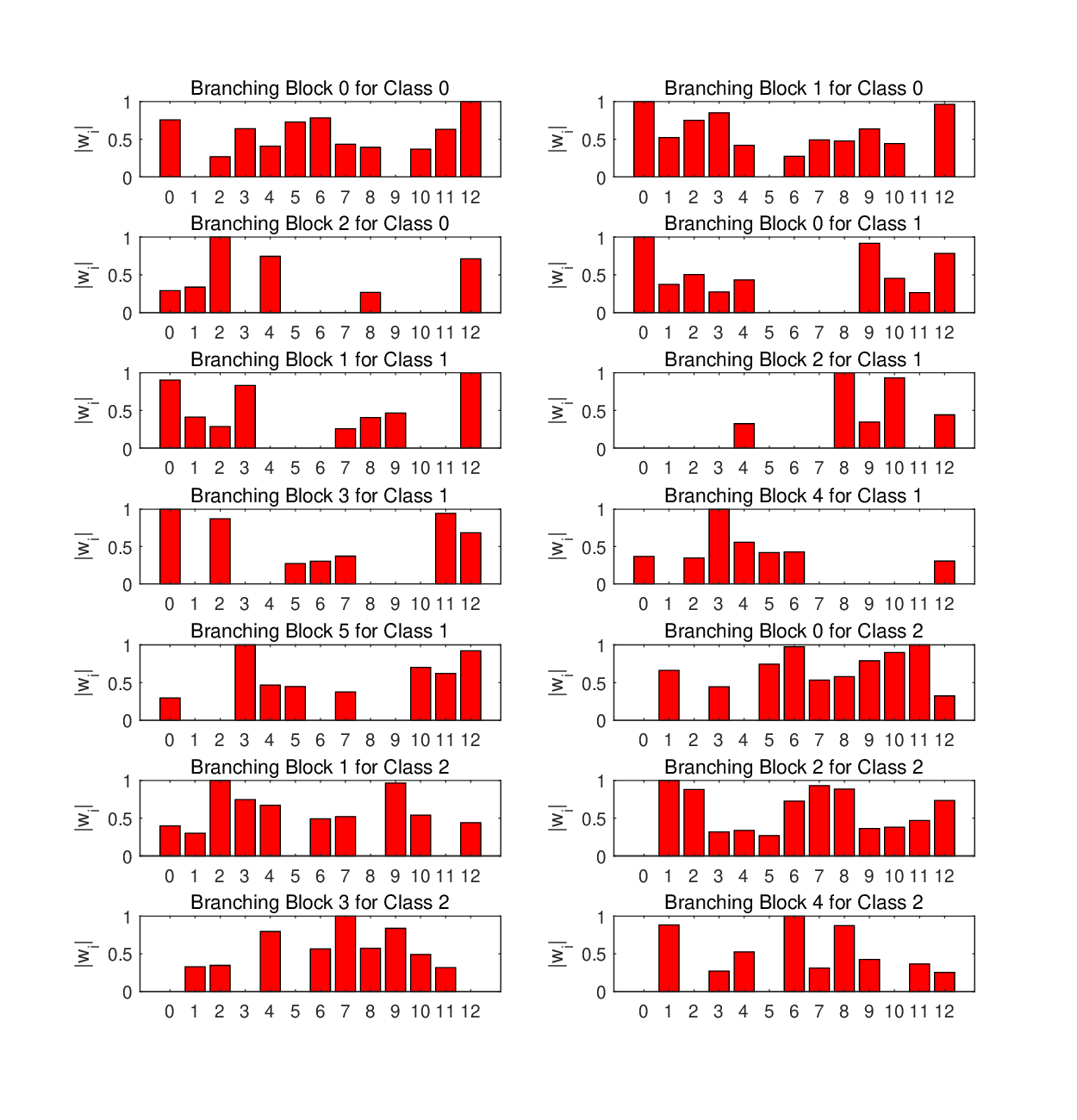}}
\caption{Visualization of the feature weights for each branching block of the three LHTs corresponding to the three classes in the wine dataset ($\beta=0.25$).}
\label{LHT_02514}
\end{center}
\vskip -0.2in
\end{figure*}

\end{document}